\documentclass{article}
\usepackage{ijcai21}

\usepackage{times}
\usepackage{soul}
\usepackage{url}
\usepackage[hidelinks]{hyperref}
\usepackage[utf8]{inputenc}
\usepackage[small]{caption}
\usepackage{graphicx}
\usepackage{amsmath}
\usepackage{amsfonts}
\usepackage{amsthm}
\usepackage{booktabs}
\usepackage{algorithm}
\usepackage{algorithmic}
\usepackage{subfigure}
\usepackage{multirow}
\usepackage{dsfont}
\usepackage{makecell}
\newtheorem{theorem}{Theorem}

\title{GraphMI: Extracting Private Graph Data from Graph Neural Networks}

\author{
Zaixi Zhang$^1$\and
Qi Liu$^1$\footnote{Contact Author}\and
Zhenya Huang$^1$\and
Hao Wang$^1$\and
Chengqiang Lu$^2$\and\\
Chuanren Liu$^3$\and
Enhong Chen$^1$\\
\affiliations
$^1$Anhui Province Key Lab. of Big Data Analysis and Application, \\School of Computer Science and Technology, University of Science and Technology of China\\
$^2$Alibaba Group\\
$^3$The University of Tennessee Knoxville\\
\emails
\{zaixi, huangzhy, wanghao3\}@mail.ustc.edu.cn,
\{qiliuql, cheneh\}@ustc.edu.cn,
lulu.lcq@alibaba-inc.com, cliu89@utk.edu
}

\begin{document}

\maketitle

\begin{abstract}
As machine learning becomes more widely used for critical applications, the need to study its implications in privacy turns to be urgent. 
Given access to the target model and auxiliary information, model inversion attack aims to infer sensitive features of the training dataset, which leads to great privacy concerns.
Despite its success in grid-like domains, directly applying model inversion techniques on non-grid domains such as graph achieves poor attack performance due to the difficulty to fully exploit the intrinsic properties of graphs and attributes of nodes used in Graph Neural Networks (GNN).
To bridge this gap, we present \textbf{Graph} \textbf{M}odel \textbf{I}nversion attack (GraphMI), which aims to extract private graph data of the training graph by inverting GNN, one of the state-of-the-art graph analysis tools.
Specifically, we firstly propose a projected gradient module to tackle the discreteness of graph edges while preserving sparsity and smoothness of graph features.
Then we design a graph auto-encoder module to efficiently exploit graph topology, node attributes, and target model parameters for edge inference.
With the proposed methods, we study the connection between model inversion risk and edge influence and show that edges with greater influence are more likely to be recovered.
Extensive experiments over several public datasets demonstrate the effectiveness of our method.
We also show that differential privacy in its canonical form can hardly defend our attack while preserving decent utility. 
\end{abstract}

\section{Introduction}

Machine learning (ML) algorithms based on deep neural networks have achieved remarkable success in a range of domains such as computer vision~\cite{zhang2020secret}, natural language processing~\cite{ijcai2020-525}, and graph analysis~\cite{wang2019mcne}.
Meanwhile, the impact of machine learning techniques on privacy is receiving more and more attention because many machine learning applications involve processing sensitive user data (e.g., purchase records)~\cite{ijcai2020-481}.
Attackers may exploit the output (i.e., black-box attack) or the parameters (i.e., white-box attack) of machine learning models to potentially reveal sensitive information in training data. 
\par
According to the attacker's goal, privacy attacks can be categorized into several types, such as membership inference attack \cite{shokri2017membership}, model extraction attack \cite{tramer2016stealing}, and model inversion attack \cite{fredrikson2015model}.
Of particular interest to this paper is model inversion attack which aims to extract sensitive features of training data given output labels and partial knowledge of non-sensitive features.
Model inversion attack was firstly introduced by \cite{fredrikson2014privacy}, where an attacker, given a linear regression model for personalized medicine and some demographic information about a patient, could predict the patient’s genetic markers.
Generally, model inversion relies on the correlation between features and output labels and try to maximize a posteriori (MAP) or likelihood estimation (MLE) to recover sensitive features.
Recently, efforts have been made to extend model inversion to attack other machine learning models, in particular Convolutional Neural Networks (CNN) ~\cite{fredrikson2015model,aivodji2019gamin}. Thus far, most model inversion attacks are investigated in the grid-like domain (e.g., images), leaving its effect on the non-grid domain (e.g., graph structured data) an open problem.

Graph Neural Network (GNN) as one of the state-of-the-art graph analysis tools shows excellent results in various applications on graph-structured data \cite{kipf2016semi,velivckovic2017graph}.  
However, the fact that many GNN-based applications such as recommendation systems \cite{wu2019session} and social relationship analysis \cite{wang2019mcne} rely on processing sensitive graph data raises great privacy concerns.
Studying model inversion attack on GNNs helps us understand the vulnerability of GNN models and enable us to avoid privacy risks in advance.
\begin{figure}[!t]
	\centering
	\includegraphics[width=0.48\textwidth]{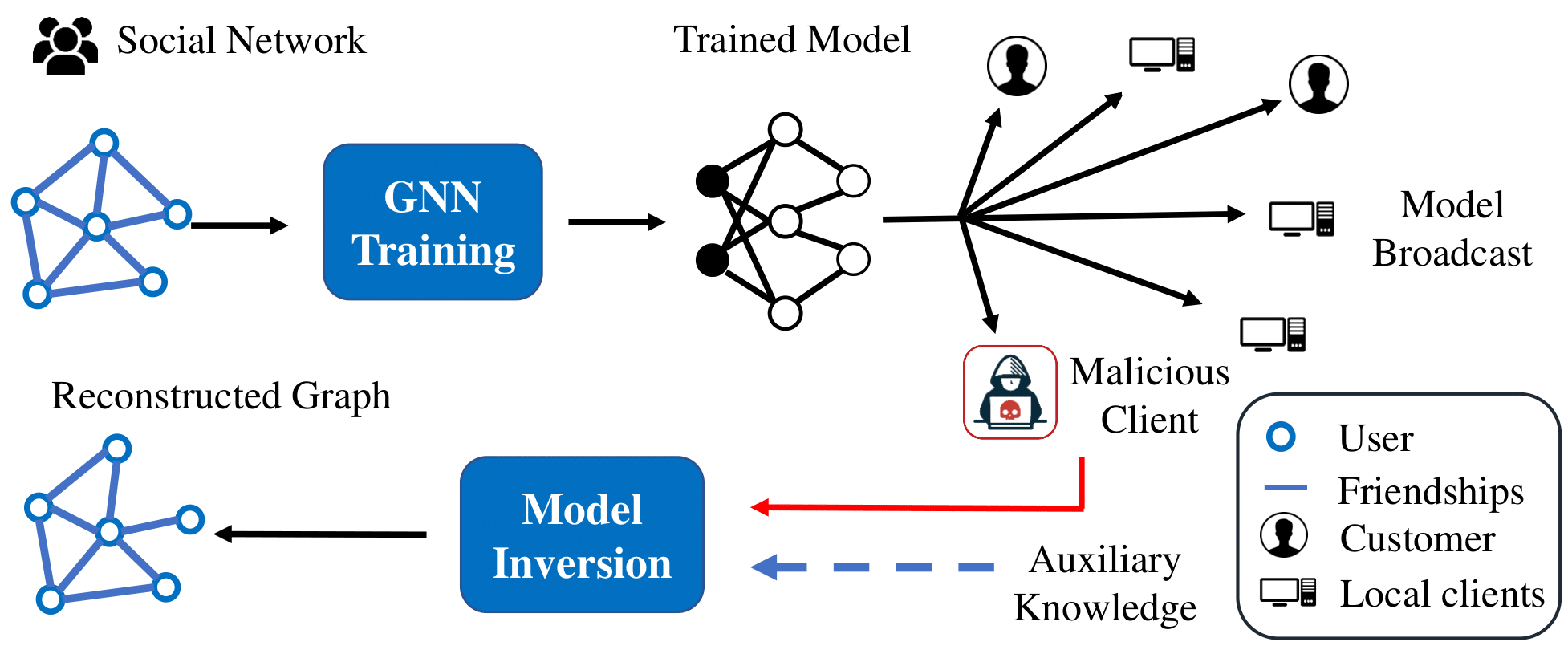}
	\caption{One motivation scenario in social networks}
	\label{motivation}
\end{figure}

\textbf{Motivation scenario:}
Figure \ref{motivation} shows one concrete motivation scenario of model inversion attack on GNN models. Users'  friendships are sensitive relational data, and users want to keep them private. Sometimes social network data is collected with user permission to train GNN models for better service. For example, these trained GNN models are used to classify friends or recommend advertisements. Then, the trained models are broadcast to customers or local clients. If the attacker can obtain the trained GNN model from malicious clients, with some auxiliary knowledge crawled from the internet, model inversion attack can be performed to reconstruct friendships among users.

In this paper, we draw attention to model inversion attacks extracting private graph data from GNN models. We focus on the following adversarial scenario. Given the trained GNN model and some auxiliary knowledge (node labels and attributes), the adversary aims to reconstruct all the edges among nodes in the training dataset. However, model inversion attack on graphs brings unique challenges. Firstly, existing model inversion attack methods can barely be applied to the graph setting due to the discrete nature of graphs. Different from the continuous image data, gradient computation and optimization on binary edges of the graph are difficult.
Secondly, current model inversion methods fail to exploit the intrinsic properties of graph such as sparsity and feature smoothness. In addition, existing model inversion attack methods cannot fully leverage the information of node attributes and GNN models. For example, node pairs with similar attributes or embeddings are more likely to have edges.

To address the aforementioned challenges, we propose \textbf{Graph} \textbf{M}odel \textbf{I}nversion attack (GraphMI) for edge reconstruction.
GraphMI is designed with two important modules: the projected gradient module and the graph auto-encoder module.
The projected gradient module is able to tackle the edge discreteness via convex relaxation while preserving graph sparsity and feature smoothness.
The graph auto-encoder module is designed to take all the information of node attributes, graph topology and target model parameters into consideration for graph reconstruction.
Based on GraphMI, we investigate the relation between edge influence and model inversion risk and find that edges with greater influence are more likely to be reconstructed.
Furthermore, we show that differential privacy, in its canonical form, is of little avail to defend against GraphMI.
Experimental results on several public datasets show the effectiveness of GraphMI \footnote{https://github.com/zaixizhang/GraphMI}.

\section{Related Work}
Based on the attacker's goal, privacy attacks can be categorized into several types such as membership inference attack \cite{shokri2017membership}, model extraction attack \cite{tramer2016stealing} and model inversion attack \cite{fredrikson2015model}. Membership inference attack tries to determine whether one sample was used to train the machine learning model; Model extraction attack is one black-box privacy attack. It tries to extract information of model parameters and reconstruct one substitute model that behaves similarly to the target model. Model inversion attack, which is the focus of this paper, aims to reconstruct sensitive features corresponding to labels of target machine learning models.\par
Model inversion attack was firstly presented in \cite{fredrikson2014privacy} for linear regression models.  \cite{fredrikson2015model} extended model inversion attack to extract faces from shallow neural networks. They cast the model inversion as an optimization problem and solve the problem by gradient descent with modifications to the images. Furthermore, several model inversion attacks in the black-box setting or assisted with generative methods are proposed~\cite{aivodji2019gamin,zhang2020secret} in the image domain. 
Thus far, no existing model inversion attack has focused on the graph domain.
\par
\section{Problem Formulation}
\subsection{Preliminaries on GNNs}
One task that GNN models are commonly used for is semi-supervised node classification~\cite{kipf2016semi}. Given  a  single  network  topology  with node attributes and a known subset of node labels, GNNs are efficient  to  infer  the  classes  of  unlabeled  nodes. Before defining GNN, we firstly introduce the following notations of graph. Let $\mathcal{G}=(\mathcal{V,E})$ be an undirected and unweighted graph, where $\mathcal{V}$ is the vertex (i.e. node) set with size $|\mathcal{V}|=N$, and $\mathcal{E}$ is the edge set. Denote $\mathcal{A}\in\{0,1\}^{N\times N}$ as an adjacent matrix containing information of network topology and $X \in \mathds{R}^{ N\times l}$ as a feature matrix with dimension $l$.  In a GNN model, each node $i$ is associated with a feature vector $\textbf{x}_i \in \mathds{R}^l$ and a scalar label $y_i$. GNN is used to predict the classes of unlabeled nodes under the adjacency matrix A and the labeled node data $\{(\textbf{x}_i,y_i)\}^{N_{train}}_{i=1}$. GNN uses all nodes' input features but only $N_{train}< N$ labeled nodes in the training phase.\par
Formally, the k-th layer of a GNN model obeys the message passing rule and can be modeled by one message passing phase and one readout update phase:
\begin{equation}
    m_v^{k+1}= \mathop{\sum}_{u\in \mathcal{N}(v)}\mathbf{M}_k(h_v^k, h_u^k, e_{uv}),
\end{equation}
\begin{equation}
    h_v^{k+1}= \mathbf{U}_k(h_v^k, m_v^{k+1}),
\end{equation}
where $\mathbf{M}_k$ denotes the message passing function and $\mathbf{U}_k$ is the vertex update function. $\mathcal{N}(v)$ is the neighbors of $v$ in graph $\mathcal{G}$. $h_v^k $ is the feature vector of node $v$ at layer $k$ and $ e_{uv}$ denotes the edge feature. $h^0_v = \textbf{x}_v $ is the input feature vector of node $v$. \par
Specifically, Graph Convolutional Network (GCN)~\cite{kipf2016semi}, a well-established method for semi-supervised node classification, obeys the following rule to aggregate neighboring features:
\begin{equation}
    H^{k+1}= \sigma \big (\hat D^{-\frac{1}{2}} \hat A \hat D^{-\frac{1}{2}} H^k W^k \big),
\end{equation}
where $\hat A = A + I_N$ is the adjacency matrix of the graph $\mathcal{G}$ with self connections added and $\hat D$ is a diagonal matrix with $\hat D_{ii} = \sum_j \hat A_{ij}$. $\sigma(\cdot)$ is the ReLU function. $H^k$ and $W^k$ are the feature matrix and the trainable weight matrix of the k-th layer respectively. $H^0 = X$ is the input feature matrix. Note that in most of this paper, we focus on two-layer GCN for the node classification. Later, we show that our graph model inversion attack can be also performed on other types of GNNs, including GAT~\cite{velivckovic2017graph} and GraphSAGE~\cite{hamilton2017inductive}.

\subsection{Problem Definition}
\begin{figure}[!t]
	\centering
	\includegraphics[width=0.48\textwidth]{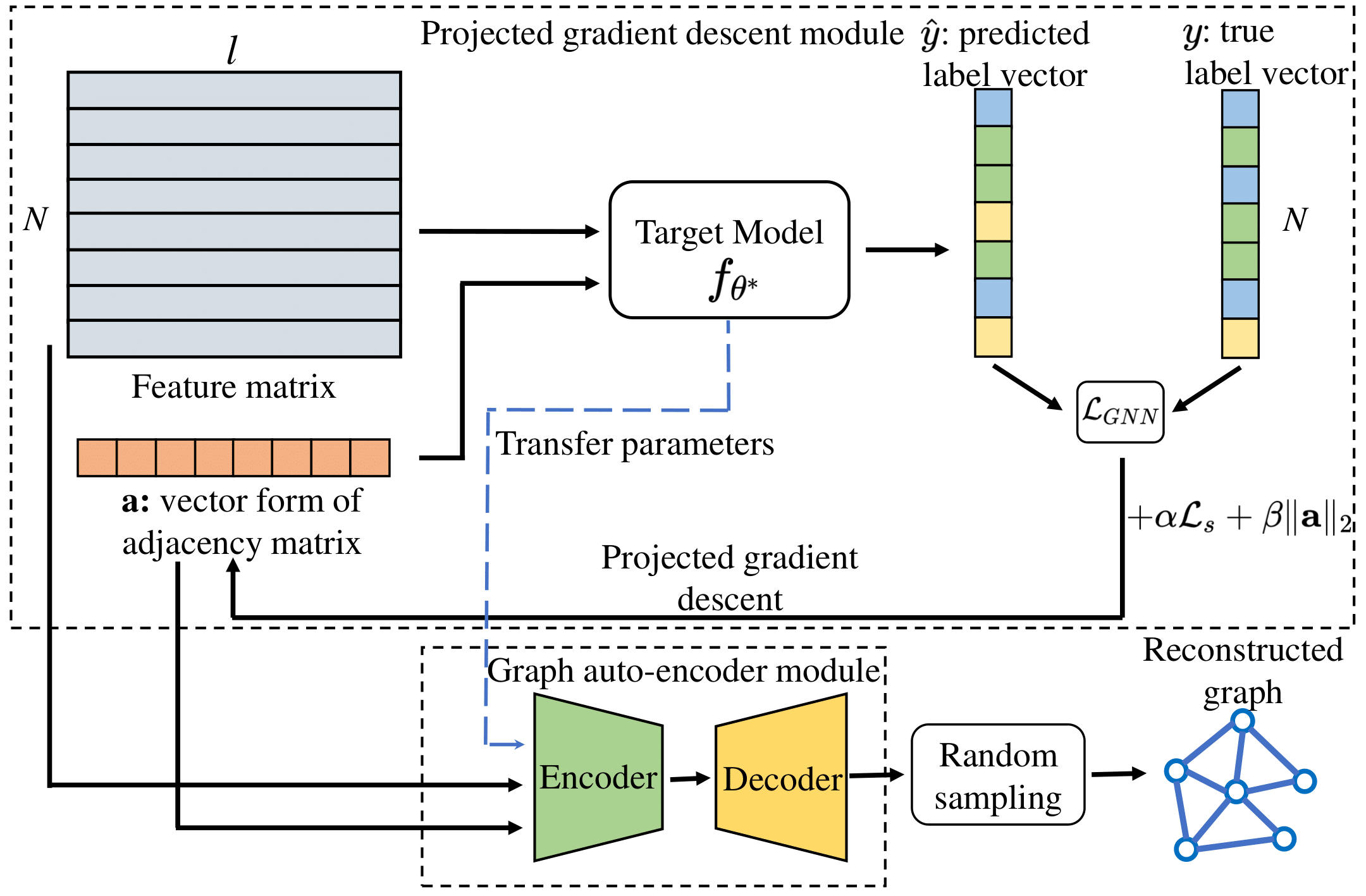}
	\caption{Overview of GraphMI}
	\label{illustration}
\end{figure}
We refer to the trained model subjected to model inversion attack as the target model. In this paper, we will firstly train a GNN for node classification task from scratch as the target model. We assume a threat model similar to the existing model inversion attacks \cite{fredrikson2015model}.

\paragraph{Attacker's Knowledge and Capability:}
We will focus on the white-box setting. The attacker is assumed to have access to the target model $f$ and can employ some inference technique to discover the adjacency matrix $A$ of the training graph. In most of the paper, we assume the attacker has labels of all the nodes. In addition to the target model $f$ and node labels, the attacker may have other auxiliary knowledge to facilitate model inversion such as node attributes, node IDs or edge density. We will discuss the impact of auxiliary knowledge and the number of node labels on attack performance in the following sections.
\paragraph{Model Inversion of Graph Neural Networks:}
Let $\theta$ be the model parameters of target model $f$. During the training phase, $f$ is trained to minimize the loss function $\mathcal{L}(\theta, X, A, Y)$:
\begin{equation}
    \theta^*={\textup{arg }}\mathop{\textup{min}}\limits_{\theta}\mathcal{L}(\theta, X, A, Y),
\end{equation}
where $Y$ is the vector of node labels and $X$ is the feature matrix.
Given the trained model and its parameters, graph model inversion aims to find the adjacency matrix $A^*$ that maximizes the posterior possibility:
\begin{equation}
    A^*={\textup{arg }}\mathop{\textup{max}}\limits_{A}P(A | X, Y, \theta^*).
\end{equation}

\begin{algorithm}[tb]
\caption{GraphMI}
\label{code}
\textbf{Input}: Target GNN model $f_{\theta*}$;
Node label vector $Y$;
Node feature matrix $X$;
Learning rate, $\eta_t$;
Iterations $T$;\\
\textbf{Output}: Reconstructed $A$
\begin{algorithmic}[1] 
\STATE $\mathbf{a}^{(0)}$ is set to zeros
\STATE Let $t=0$
\WHILE{t \textless T}
\STATE Gradient descent: $\mathbf{a}^{(t)}= \mathbf{a}^{(t-1)}- \eta_t \nabla \mathcal{L}_{attack}(\mathbf{a})$;
\STATE Call Projection operation in (\ref{projection})
\ENDWHILE
\STATE Call Graph auto-encoder module in (\ref{GAE})
\STATE Call Random sampling module.
\STATE \textbf{return} $A$
\end{algorithmic}
\end{algorithm}

\section{Proposed Algorithm}
Next we introduce GraphMI, our proposed model inversion attack on GNN models.
\subsection{Attack Overview}
Figure \ref{illustration} shows the overview of GraphMI. Generally, GraphMI is one optimization-based attack method, which firstly employs projected gradient descent on the graph to find the ``optimal'' network topology for node labels. Then the adjacency matrix and feature matrix will be sent to the graph auto-encoder module of which parameters are transfered from the target model. Finally, we can interpret the optimized graph as the edge probability matrix and sample a binary adjacency matrix. We summarize GraphMI in Algo \ref{code}.
\subsection{Details of Modules}
\paragraph{Projected Gradient Descent Module:} We treat model inversion on GNNs as one optimization problem: given node features or node IDs, we want to minimize the cross-entropy loss between true labels $y_i$ and predicted labels $\hat{y_i}$ from the target GNN model $f_{\theta*}$. The intuition is that the reconstructed adjacency matrix will be similar to the original adjacency matrix if the loss between true labels and predicted labels is minimized. The attack loss on node $i$ is denoted by $\ell_i(A, f_{\theta^*}, \textbf{x}_i, y_i)$ where $A$ is the reconstructed adjacency matrix, $\theta^*$ is the model parameter of the target model $f$ and $\textbf{x}_i$ is the node feature vector of node $i$. The objective function can be formulated as:
\begin{equation}
    \begin{split}
    \mathop{\textup{min}}\limits_{A \in \{0,1\}^{N\times N}} \mathcal{L}_{GNN}(A) &=\frac{1}{N}\sum_{i=1} ^N \ell_i(A, f_{\theta^*}, \textbf{x}_i, {y_i}) \\s.t.\quad A &= A^\top.
\end{split}
\label{equ1}
\end{equation}
\par
In many real-world graphs, such as social networks, citation networks, and web pages, connected nodes are likely to have similar features \cite{wu2019adversarial}. Based on this observation, we need to ensure the feature smoothness in the optimized graph. The feature smoothness can be captured by the following loss term $\mathcal{L}_s$:
\begin{equation}
\begin{split}
    \mathcal{L}_{s} = \frac{1}{2}\sum_{i, j =1} ^N A_{i, j}(\textbf{x}_i-\textbf{x}_j)^2,
\end{split}
\end{equation}
where $A_{i,j}$ indicates the connection between node $v_i$ and $v_j$ in the optimized graph and $(\textbf{x}_i-\textbf{x}_j)^2$ measures the feature difference between $v_i$ and $v_j$. $\mathcal{L}_s$ can also be represented as:
\begin{equation}
\begin{split}
    \mathcal{L}_s = tr(X^\top L X),
\end{split}
\end{equation}
where $L=D-A$ is the laplacian matrix of $A$ and $D$ is the diagonal matrix of $A$. In this paper, to make feature smoothness
independent of node degrees, we use the normalized lapacian matrix $\hat{L} = D^{-1/2}LD^{-1/2}$ instead: 
\begin{equation}
\begin{split}
    \mathcal{L}_s = tr(X^\top \hat{L} X)=\frac{1}{2}\sum_{i, j =1} ^N A_{i, j}(\frac{\textbf{x}_i}{\sqrt{d_i}}-\frac{\textbf{x}_j}{\sqrt{d_j}})^2,
\end{split}
\end{equation}
where $d_i$ and $d_j$ denote the degree of node $v_i$ and $v_j$. To encourage the sparsity of graph structure, F norm of adjacency matrix $A$ is also added to the loss function. The final objective function is:
\begin{equation}
\begin{split}
    {\textup{arg }}\mathop{\textup{min}}\limits_{A \in \{0,1\}^{N\times N}}\mathcal{L}_{attack} &=\mathcal{L}_{GNN}+\alpha \mathcal{L}_s+\beta \|A\|_F
    \\s.t. \quad A &= A^\top,\label{equ3}
\end{split}
\end{equation}
where $\alpha$ and $\beta$ are hyper-parameters that control the contribution of feature smoothing and graph sparsity.
Solving equation (\ref{equ3}) is a combinatorial optimization problem due to edge discreteness. For ease of gradient computation and update, we firstly replace the symmetric reconstructed adjacency matrix $A$ with its vector form $\mathbf{a}$ that consists of $n:=N(N-1)/2$ unique variables in $A$. Adjacency matrix $A$ and vector $\mathbf{a}$ can be converted to each other easily, which ensures the optimized adjacency matrix is symmetric. Then we relax $\mathbf{a} \in \{0,1\}^n$ into convex space $\mathbf{a} \in [0,1]^n$. We can perform model inversion attack by firstly solving the following optimization problem:
\begin{equation}
    {\textup{arg }}\mathop{\textup{min}}\limits_{\mathbf{a} \in [0,1]^n}\mathcal{L}_{attack} =\mathcal{L}_{GNN}+\alpha \mathcal{L}_s+\beta \|\mathbf{a}\|_2.
\label{equ4}
\end{equation}

\par The continuous optimization problem \ref{equ4} is solved by projected gradient descent (PGD):
\begin{equation}
    \mathbf{a}^{t+1} = P_{[0,1]} \big[\mathbf{a}^t -\eta_t g_t \big],
\end{equation}
where $t$ is the iteration index of PGD, $\eta_t$ is the learning rate, $g_t$ is the gradients of loss $\mathcal{L}_{attack}$ in \ref{equ3} evaluated at $\textbf{a}^t$, and
\begin{equation}
P_{[0,1]}[x]=\left\{
\begin{array}{rcl}
0 & & x < 0\\
1 & & x > 1\\
x & & otherwise
\end{array} \right.
\label{projection}
\end{equation}
is the projection operator.
\paragraph{Graph Auto-encoder Module:} In GraphMI, we propose to use graph auto-encoder (GAE) \cite{kipf2016variational} to post-process the optimized adjacency matrix $A$. GAE is composed of two components: encoder and decoder. We transfer part of the parameters from the target model $f_{\theta^*}$ to the encoder. Specifically, feature matrix and adjacency vector $\mathbf{a}$ are sent to the $f_{\theta^*}$ and the node embedding matrix $Z$ is generated by taking the penultimate layer of the target model $f_{\theta^*}$, which is denoted as $H_{\theta^*}(\mathbf{a}, X)$. Then the decoder will reconstruct adjacency matrix $A$ by applying logistic sigmoid function to the inner product of $Z$:
\begin{equation}
    A={\rm sigmoid}(ZZ^\top),  {\rm with}~ Z=H_{\theta^*}(\mathbf{a}, X).\label{GAE}
\end{equation}
\par The node embeddings generated by the graph auto-encoder module encode the information from node attributes, graph topology, and the target GNN model. Intuitively, node pairs with close embeddings are more likely to form edges.
\paragraph{Random Sampling Module:} After the optimization problem is solved, the solution $A$ can be interpreted as a probabilistic matrix, which represents the possibility of each edge. We could use random sampling to recover the binary adjacency matrix; see details in the appendix.\\

\subsection{Analysis on Correlation between Edge Influence and Inversion Risk}
In previous work \cite{wu2016methodology}, researchers found feature influence to be an essential factor in incurring privacy risk.  In our context of graph model inversion attack, sensitive features are edges. Here we want to characterize the correlation between edge influence and inversion risk. Given label vector $Y$, adjacency matrix $A$ and feature matrix $X$, the performance of target model $f_{\theta^*}$ for the prediction can be measured by prediction accuracy:
\begin{equation}
    ACC(f_{\theta^*}, A, X)=\frac{1}{N}\sum_{i=1}^N\mathds{1}(f_{\theta^*}^i (A, X)=y_i),
\end{equation}
where, $f_{\theta^*}^i (A, X)$ is the predicted label for node i. The influence of edge $e$ can be defined as:
\begin{equation}
     {\mathcal I}(e)=ACC(f_{\theta^*}, A, X) - ACC(f_{\theta^*}, A_{-e}, X),
     \label{edge}
\end{equation} 
where $A_{-e}$ denotes removing the edge $e$ from the adjacency matrix $A$. \cite{wu2016methodology} proposed to use adversary advantage to characterize model inversion risk of features. The model inversion advantage of adversary $\mathcal{A}$ is defined to be $P[\mathcal{A}(X, f_{\theta^*})=e]- 1/2$, where $P[\mathcal{A}(X, f_{\theta^*})= e]$ is the probability that adversary $\mathcal{A}$ correctly infer the existence of edge $e$. Next, we introduce our theorem.
\begin{theorem}
The adversary advantage is greater for edges with greater influence.
\end{theorem}
We defer the proof to the appendix. Intuitively, edges with greater influence are more likely to be recovered by GraphMI because these edges have a greater correlation with the model output. In the following section, we will validate our theorem with experiments.

\section{Experiments}
\begin{table*}[!t]
\centering
\begin{tabular}{lp{0.7cm}<{\centering}p{0.7cm}<{\centering}p{0.7cm}<{\centering}p{0.7cm}<{\centering}p{0.7cm}<{\centering}p{0.7cm}<{\centering}p{0.7cm}<{\centering}p{0.7cm}<{\centering}p{0.7cm}<{\centering}p{0.7cm}<{\centering}p{0.7cm}<{\centering}p{0.7cm}<{\centering}p{0.7cm}<{\centering}p{0.7cm}<{\centering}}
\toprule
                \multirow{2}{*}{\textbf{Method}}& \multicolumn{2}{c}{\textbf{Cora}}                         & \multicolumn{2}{c}{\textbf{Citeseer}}                     & \multicolumn{2}{c}{\textbf{Polblogs}}& \multicolumn{2}{c}{\textbf{USA}} &\multicolumn{2}{c}{\textbf{Brazil}} &\multicolumn{2}{c}{\textbf{AIDS}} &\multicolumn{2}{c}{\textbf{ENZYMES}} \\ 
                \cmidrule(r){2-3} \cmidrule(r){4-5} \cmidrule(r){6-7} \cmidrule(r){8-9}\cmidrule(r){10-11}\cmidrule(r){12-13}\cmidrule(r){14-15}
                         & AUC           & AP            & AUC      & AP            & AUC            & AP      & AUC &AP&AUC&AP&AUC &AP&AUC&AP  \\ \midrule
Attr. Sim. & 0.803          & 0.808          & \textbf{0.889}          & \textbf{0.891}&-&-&-&-&-&-          & 0.731          & 0.727          & 0.564                       & 0.567 \\ 
MAP   & 0.747          & 0.708          & 0.693          & 0.755          & 0.688          & 0.751          & 0.594                       & 0.601                       & 0.638 &0.661&0.642&0.653&0.617&0.643                             \\ 
GraphMI     & \textbf{0.868} & \textbf{0.883} & 0.878 & 0.885 & \textbf{0.793} & \textbf{0.797} & \textbf{0.806}& \textbf{0.813}  & \textbf{0.866} &\textbf{0.888}&\textbf{0.802} &\textbf{0.809}&\textbf{0.678}&\textbf{0.684}                                                    \\ \bottomrule
\end{tabular}
\caption{Results of model inversion attack on Graph Neural Networks}
\label{result}
\end{table*}
In this section, we present the experimental results to show the effectiveness of GraphMI. Specifically, our experiments are designed to answer the following research questions:
\begin{itemize}
    \item \textbf{RQ1}: How effective is GraphMI?
    \item \textbf{RQ2}: Which edges are more likely to be 
    \item \textbf{RQ3}: Is differential privacy an effective countermeasure against model inversion attacks on GNN?
\end{itemize}

\subsection{Experimental Settings}

\textbf{Datasets: }Our graph model inversion attack method is evaluated on 7 public datasets from 4 categories. The detailed statistics of them are listed in the appendix.
\begin{itemize}
    \item \textit{Citation Networks}: We use Cora and Citeseer \cite{sen2008collective}. Here, nodes are documents with corresponding bag-of-words features and edges denote citations among nodes. Class labels denote the subfield of research that
    the papers belong to.
    \item \textit{Social Networks}: Polblogs \cite{adamic2005political} is the  network of political blogs whose nodes do not have features. 
    \item \textit{Air-Traffic Networks}:  The air-traffic networks are based on flight records from USA and Brazil. Each node is an airport and
    an edge indicates a commercial airline route
    between airports. Labels denote the level of activity
    in terms of people and flights passing through an
    airport \cite{ribeiro2017struc2vec}.
    \item \textit{Chemical Networks}: AIDS \cite{riesen2008iam} and ENZYMES \cite{borgwardt2005protein} are chemical datasets that contain many molecure graphs, each node is an atom and each link represents chemical bonds.
\end{itemize}

\paragraph{Target Models:}In our evaluation, we use 3 state-of-the-art GNN models: GCN \cite{kipf2016semi}, GAT \cite{velivckovic2017graph} and GraphSAGE \cite{hamilton2017inductive}. 
The parameters of the models are the same as those set in the original papers. To train a target model, 10\% randomly sampled nodes are used as the training set. All GNN models are trained for 200 epochs with an early stopping strategy based on convergence behavior and accuracy on a validation set containing 20\% randomly sampled nodes. In GraphMI attack experiments, attackers have labels of all the nodes and feature vectors. All the experiments are conducted on Tesla V100 GPUs.
\paragraph{Parameter Settings:}In experiments, we set $\alpha=0.001$, $\beta=0.0001$, $\eta_t=0.1$ and $T=100$ as the default setting. We show how to find optimal values for hyper-parameters in the following section.
\paragraph{Metrics:}Since our attack is unsupervised, the attacker cannot find a threshold to make a concrete prediction through the  algorithm. To evaluate our attack, we use AUC (area under the ROC curve) and AP (average precision) as our metrics, which is consistent with previous works \cite{kipf2016variational}. In experiments, we use all the edges from the training graph and the same number of randomly sampled pairs of unconnected nodes (non-edges) to evaluate AUC and AP.
\begin{figure}[t]
	\centering
    \includegraphics[width=0.7\linewidth]{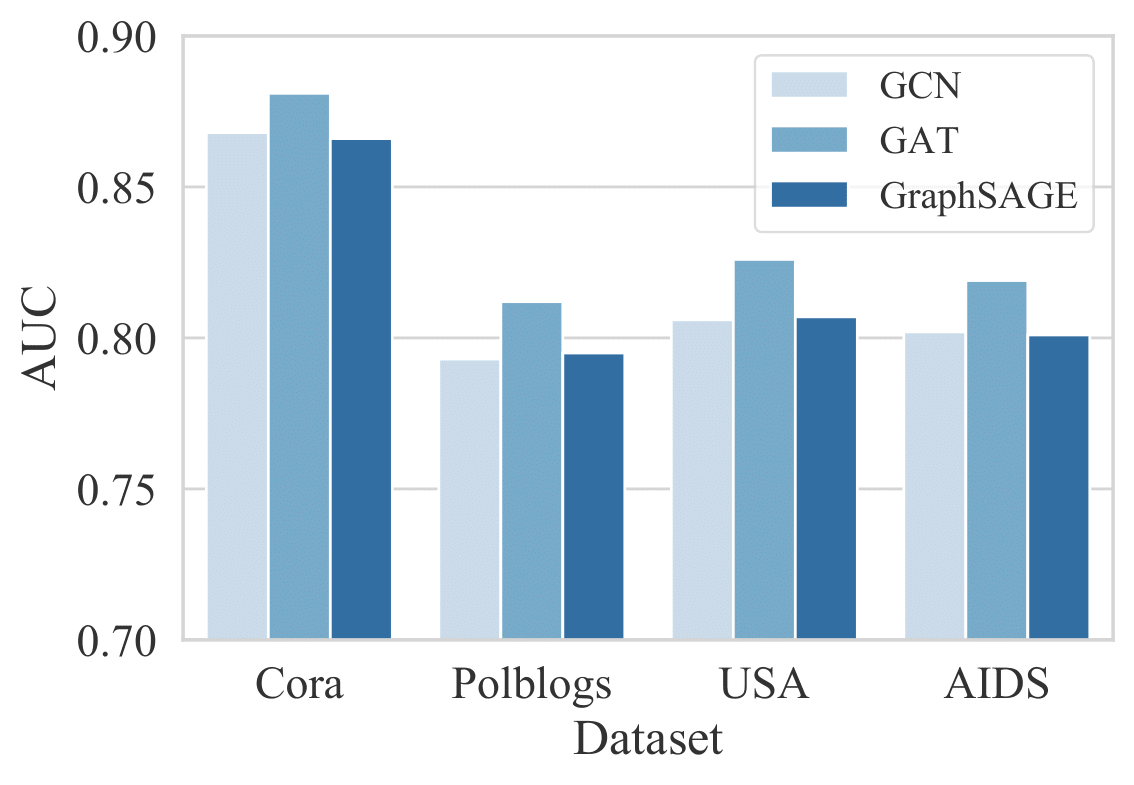}
	\caption{Attack performance of GraphMI on different Graph Neural Networks.}
	\label{gnn type}
\end{figure}
\subsection{Results and Discussions}
\paragraph{Attack Performance.}Results for model inversion attacks on GCN are summarized in table \ref{result}. There are two baseline methods, attribute similarity (abbreviated as Attr. Sim.) and MAP. Attribute similarity is measured by cosine distance among node attributes, which is commonly used in previous works \cite{he2020stealing}. We adapt the model inversion method from \cite{fredrikson2015model}, MAP to the graph neural network setting as the other baseline. Note that some datasets such as Polblogs dataset do not have node attributes, so that we assign one-hot vectors as their attributes. They are not applicable for the attack based on attribute similarity. As can be observed in table \ref{result}, GraphMI achieves the best performance across nearly all the datasets, which demonstrates the effectiveness of GraphMI. One exception is Citesser where the attack performance of GraphMI is relatively lower than attribute similarity, which could be explained by more abundant node attribute information of Citeseer compared with other datasets. Thus using node attribute similarity alone could achieve good performance in the Citesser dataset.

In figure \ref{gnn type}, we show the attack performance of GraphMI on three GNNs. We observe that GraphMI has better attack performance on GAT model. This may be explained by the fact that GAT model is more powerful and is able to build a stronger correlation between graph topology and node labels. GraphMI can take advantage of such a stronger correlation and achieve better attack performance. \par
\begin{figure}[t]
	\centering
	\includegraphics[width=0.7\linewidth]{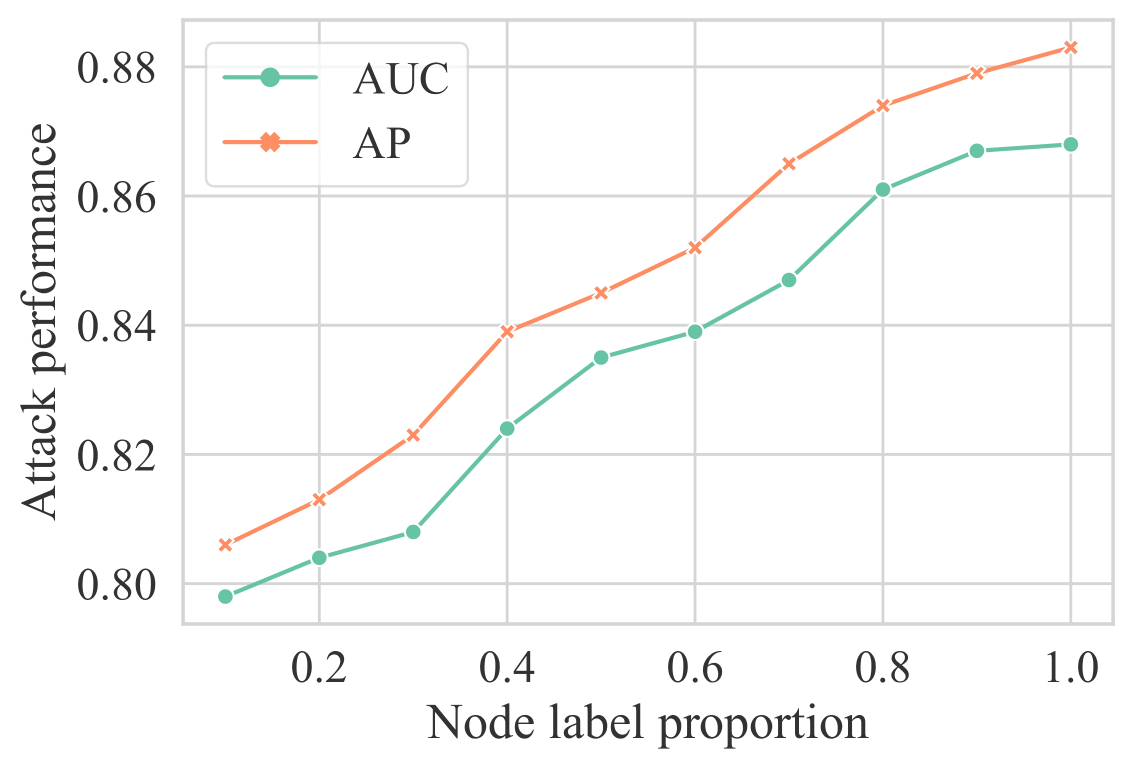}
	\caption{Impact of node label proportion.}
	\label{rq1}
\end{figure}

 
In figure \ref{rq1}, we present the influence of node label proportion on attack performance. As can be observed from the plot, with fewer node labels, the attack performance will drop gradually. But GraphMI can still achieve over 80 $\%$ AUC and AP when only 20$\%$ node labels are available, which again verifies the effectiveness of GraphMI.
\par
\begin{figure}[t]
	\centering
	\subfigure[]{\includegraphics[width=0.48\linewidth]{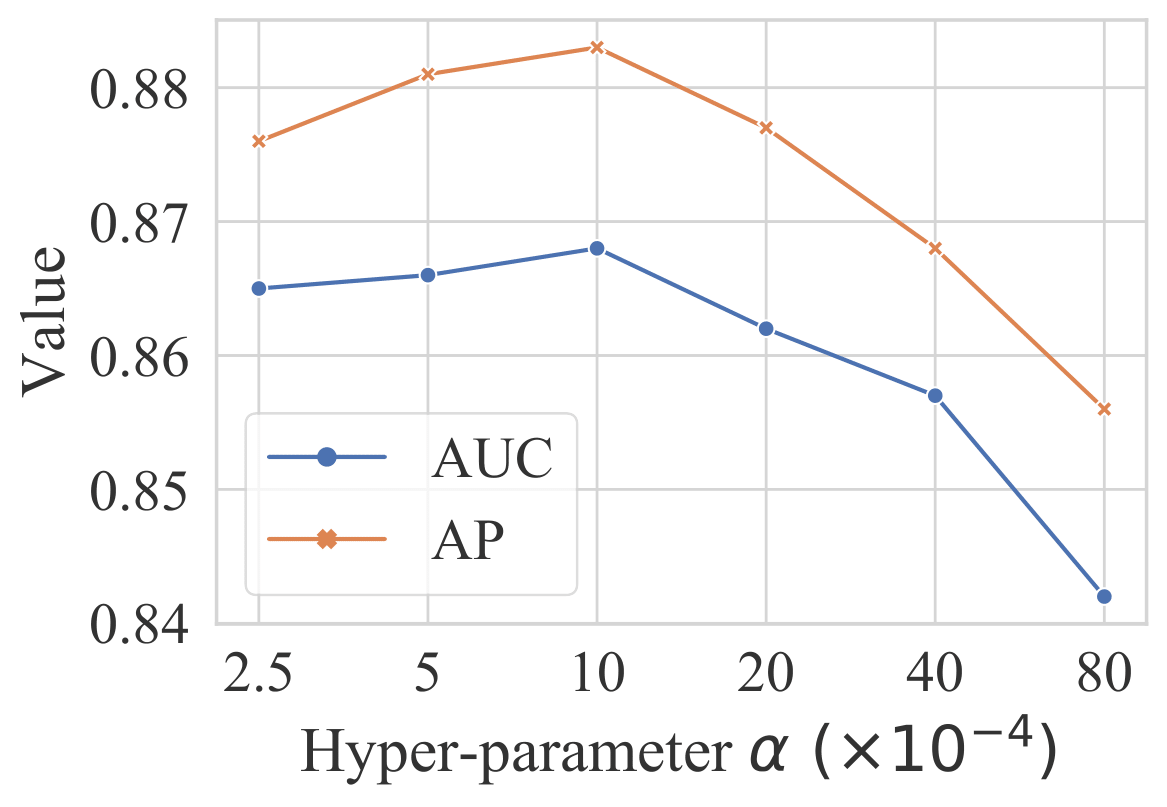}}
    \subfigure[]{\includegraphics[width=0.48\linewidth]{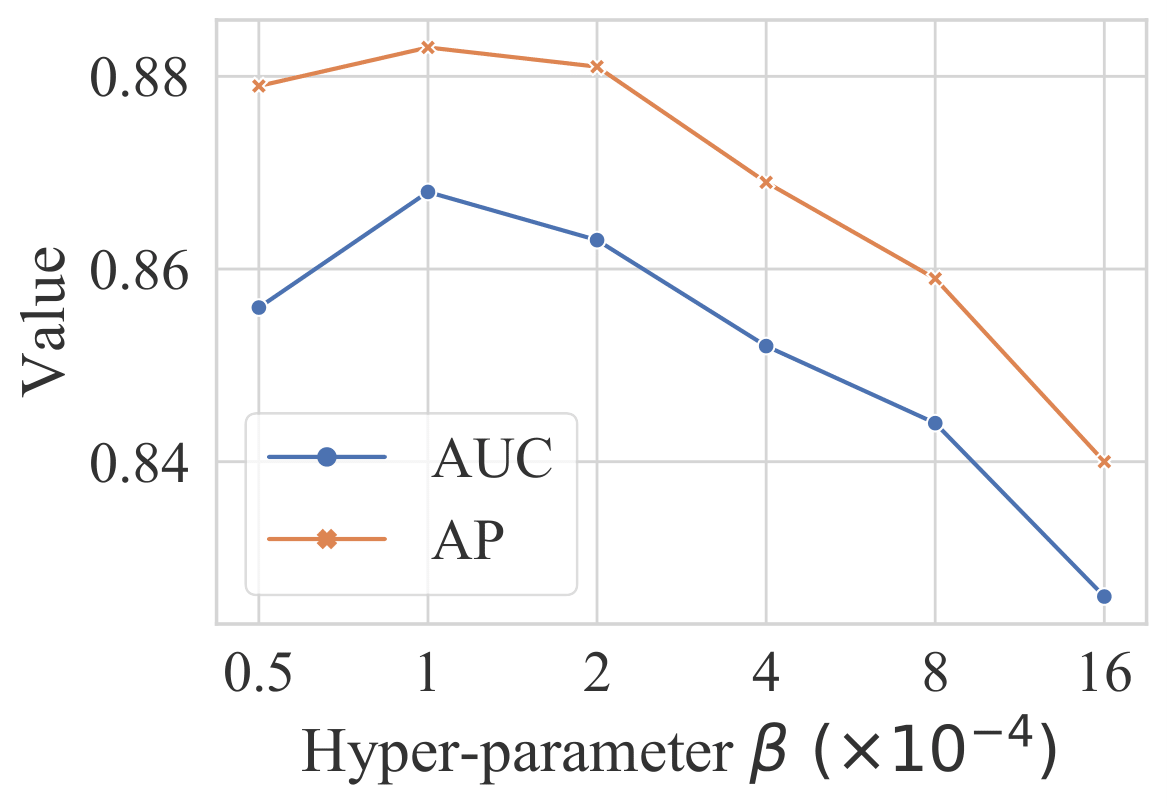}}
	\caption{ Results of parameter analysis on Cora dataset}
	\label{param analysis}
\end{figure}

 We also explore the sensitivity of hyper-parameters $\alpha$ and $\beta$ for GraphMI. In the experiments, we alter the value of
$\alpha$ and $\beta$ to see how they affect the performance of GraphMI. Specifically, we vary $\alpha$ from 0.00025 to 0.008 and $\beta$ from 0.00005 to 0.0016 in a log scale of base 2. The attack performance change of GraphMI is illustrated in Fig~\ref{param analysis}. As we can observe, the attack performance of GraphMI can be boosted when choosing proper values for all the hyper-parameters.

\begin{figure}[t]
	\centering
    \includegraphics[width=0.7\linewidth]{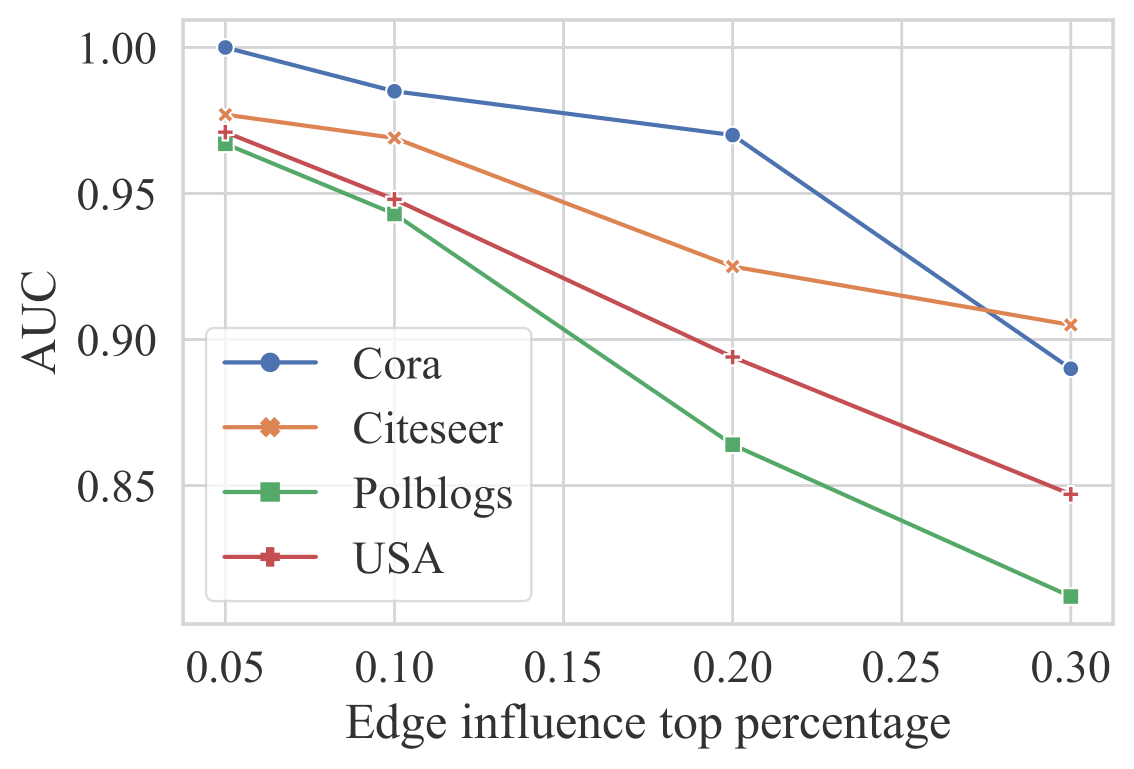}
	\caption{Impact of edge influence on the performance of the GraphMI attack.}
	\label{edge influence}
\end{figure}

\paragraph{Edge Influence.}We do experiments to verify our claim that edges with greater influence are more likely to be inferred successfully through model inversion attack. Note that it will be very time-consuming to measure the influence of each edge exactly. According to equation (\ref{edge}), removing edges with greater influence will cause greater drop of prediction accuracy. To select edges with great influence, we apply the state-of-the-art topology attack \cite{xu2019topology} on graphs by removing edges. In Figure \ref{edge influence}, we show that for edges with top $5\%$ influence GraphMI achieves the attack AUC of nearly 1.00 in Cora dataset. This implies that the privacy leakage will be more severe if sensitive edges are those with greater influence.

\begin{table}[h]
\centering
\begin{tabular}{lrc}
\toprule
Method & ACC & GraphMI AUC\\ \midrule
$\epsilon=1.0 $ &                0.48             &         0.60       \\ 
$\epsilon=5.0 $  &               0.65              &        0.72        \\ 
$\epsilon=10.0 $ &               0.78               &         0.84       \\ 
no DP   &                        0.80      &        0.87        \\ \bottomrule
\end{tabular}
\caption{The performance of the GraphMI attack against GCN trained with differential privacy on Cora dataset}
\label{DP}
\end{table}
\paragraph{Defense Performance of Differentail Privacy.}
Differential privacy (DP) is one general approach for protecting privacy.
Here, we investigate the impact of differential privacy on GraphMI attacks. $(\epsilon,\delta)$ -
DP is ensured by adding Gaussian noise to clipped gradients in each training iteration \cite{abadi2016deep}. In experiments, $\delta$ is set to $10^{-5}$ and the noise scale is varied to obtain target GNN models with different $\epsilon$ from 1.0 to 10.0.
The GraphMI attack performance and their model utility are presented in Table \ref{DP}. As the privacy budget $\epsilon$ drops, the performance of GraphMI attack deteriorates at the price of a huge utility drop. Generally, enforcing DP on target models cannot prevent GraphMI attack.

\section{Conclusion}

In this paper, we presented GraphMI, a model inversion attack method against Graph Neural Networks.
Our method was specifically designed and optimized for extracting private graph-structured data from GNNs.
Extensive experimental results showed its effectiveness on several state-of-the-art graph neural networks.
We also explored and evaluated the impact of node label proportion and edge influence on the attack performance.
Finally, we showed that imposing differential privacy on graph neural networks can hardly protect privacy while preserving decent utility.

This paper provided potential tools for investigating the privacy risks of deep learning models on graph-structured data.
Interesting future directions include: 1) Extending the current work to a black-box setting.2) Design countermeasures with a better trade-off between utility and privacy.

\section*{Acknowledgments}
This research was partially supported by grants from the National Natural Science Foundation of China (Grants No.61922073 and U20A20229), and the Fundamental Research Funds for the Central Universities (Grant No.WK2150110021).

\bibliographystyle{named}
\bibliography{paper}

\begin{thebibliography}{}

\bibitem[\protect\citeauthoryear{Abadi \bgroup \em et al.\egroup
  }{2016}]{abadi2016deep}
Martin Abadi, Andy Chu, Ian Goodfellow, H~Brendan McMahan, Ilya Mironov, Kunal
  Talwar, and Li~Zhang.
\newblock Deep learning with differential privacy.
\newblock In {\em CCS 2016}, pages 308--318, 2016.

\bibitem[\protect\citeauthoryear{Adamic and Glance}{2005}]{adamic2005political}
Lada~A Adamic and Natalie Glance.
\newblock The political blogosphere and the 2004 us election: divided they
  blog.
\newblock In {\em Proceedings of the 3rd international workshop on Link
  discovery}, pages 36--43, 2005.

\bibitem[\protect\citeauthoryear{A{\"\i}vodji \bgroup \em et al.\egroup
  }{2019}]{aivodji2019gamin}
Ulrich A{\"\i}vodji, S{\'e}bastien Gambs, and Timon Ther.
\newblock Gamin: An adversarial approach to black-box model inversion.
\newblock {\em arXiv preprint arXiv:1909.11835}, 2019.

\bibitem[\protect\citeauthoryear{Borgwardt \bgroup \em et al.\egroup
  }{2005}]{borgwardt2005protein}
Karsten~M Borgwardt, Cheng~Soon Ong, Stefan Sch{\"o}nauer, SVN Vishwanathan,
  Alex~J Smola, and Hans-Peter Kriegel.
\newblock Protein function prediction via graph kernels.
\newblock {\em Bioinformatics}, 21(suppl\_1):i47--i56, 2005.

\bibitem[\protect\citeauthoryear{Fioretto \bgroup \em et al.\egroup
  }{2020}]{ijcai2020-481}
Ferdinando Fioretto, Lesia Mitridati, and Pascal Van~Hentenryck.
\newblock Differential privacy for stackelberg games.
\newblock In {\em IJCAI-20}, pages 3480--3486, 7 2020.
\newblock Main track.

\bibitem[\protect\citeauthoryear{Fredrikson \bgroup \em et al.\egroup
  }{2014}]{fredrikson2014privacy}
Matthew Fredrikson, Eric Lantz, Somesh Jha, Simon Lin, David Page, and Thomas
  Ristenpart.
\newblock Privacy in pharmacogenetics: An end-to-end case study of personalized
  warfarin dosing.
\newblock In {\em 23rd $\{$USENIX$\}$ Security Symposium}, pages 17--32, 2014.

\bibitem[\protect\citeauthoryear{Fredrikson \bgroup \em et al.\egroup
  }{2015}]{fredrikson2015model}
Matt Fredrikson, Somesh Jha, and Thomas Ristenpart.
\newblock Model inversion attacks that exploit confidence information and basic
  countermeasures.
\newblock In {\em SIGSAC}, pages 1322--1333, 2015.

\bibitem[\protect\citeauthoryear{Hamilton \bgroup \em et al.\egroup
  }{2017}]{hamilton2017inductive}
Will Hamilton, Zhitao Ying, and Jure Leskovec.
\newblock Inductive representation learning on large graphs.
\newblock In {\em NeurIPS}, pages 1024--1034, 2017.

\bibitem[\protect\citeauthoryear{He \bgroup \em et al.\egroup
  }{2021}]{he2020stealing}
Xinlei He, Jin-Yuan Jia, M.~Backes, N.~Gong, and Y.~Zhang.
\newblock Stealing links from graph neural networks.
\newblock In {\em 30th {USENIX} Security Symposium}, Vancouver, B.C., August
  2021. {USENIX} Association.

\bibitem[\protect\citeauthoryear{Kipf and Welling}{2016}]{kipf2016variational}
Thomas~N Kipf and Max Welling.
\newblock Variational graph auto-encoders.
\newblock {\em Bayesian Deep Learning Workshop (NeurIPS 2016)}, 2016.

\bibitem[\protect\citeauthoryear{Kipf and Welling}{2017}]{kipf2016semi}
Thomas~N Kipf and Max Welling.
\newblock Semi-supervised classification with graph convolutional networks.
\newblock {\em {ICLR} 2017}, 2017.

\bibitem[\protect\citeauthoryear{Liu \bgroup \em et al.\egroup
  }{2020}]{ijcai2020-525}
Xin Liu, Kai Liu, Xiang Li, Jinsong Su, Yubin Ge, Bin Wang, and Jiebo Luo.
\newblock An iterative multi-source mutual knowledge transfer framework for
  machine reading comprehension.
\newblock In {\em IJCAI-20}, pages 3794--3800, 7 2020.

\bibitem[\protect\citeauthoryear{Ribeiro \bgroup \em et al.\egroup
  }{2017}]{ribeiro2017struc2vec}
Leonardo~FR Ribeiro, Pedro~HP Saverese, and Daniel~R Figueiredo.
\newblock struc2vec: Learning node representations from structural identity.
\newblock In {\em SIGKDD}, pages 385--394, 2017.

\bibitem[\protect\citeauthoryear{Riesen and Bunke}{2008}]{riesen2008iam}
Kaspar Riesen and Horst Bunke.
\newblock Iam graph database repository for graph based pattern recognition and
  machine learning.
\newblock In {\em Joint IAPR International Workshops on SPR and SSPR}, pages
  287--297. Springer, 2008.

\bibitem[\protect\citeauthoryear{Sen \bgroup \em et al.\egroup
  }{2008}]{sen2008collective}
Prithviraj Sen, Galileo Namata, Mustafa Bilgic, Lise Getoor, Brian Galligher,
  and Tina Eliassi-Rad.
\newblock Collective classification in network data.
\newblock {\em AI magazine}, 29(3):93--93, 2008.

\bibitem[\protect\citeauthoryear{Shokri \bgroup \em et al.\egroup
  }{2017}]{shokri2017membership}
Reza Shokri, Marco Stronati, Congzheng Song, and Vitaly Shmatikov.
\newblock Membership inference attacks against machine learning models.
\newblock In {\em 2017 IEEE Symposium on Security and Privacy (SP)}, pages
  3--18. IEEE, 2017.

\bibitem[\protect\citeauthoryear{Tram{\`e}r \bgroup \em et al.\egroup
  }{2016}]{tramer2016stealing}
Florian Tram{\`e}r, Fan Zhang, Ari Juels, Michael~K Reiter, and Thomas
  Ristenpart.
\newblock Stealing machine learning models via prediction apis.
\newblock In {\em 25th $\{$USENIX$\}$ Security Symposium}, pages 601--618,
  2016.

\bibitem[\protect\citeauthoryear{Veli{\v{c}}kovi{\'c} \bgroup \em et al.\egroup
  }{2018}]{velivckovic2017graph}
Petar Veli{\v{c}}kovi{\'c}, Guillem Cucurull, Arantxa Casanova, Adriana Romero,
  Pietro Lio, and Yoshua Bengio.
\newblock Graph attention networks.
\newblock {\em ICLR 2018}, 2018.

\bibitem[\protect\citeauthoryear{Wang \bgroup \em et al.\egroup
  }{2019}]{wang2019mcne}
Hao Wang, Tong Xu, Qi~Liu, Defu Lian, Enhong Chen, Dongfang Du, Han Wu, and Wen
  Su.
\newblock Mcne: An end-to-end framework for learning multiple conditional
  network representations of social network.
\newblock In {\em Proceedings of the 25th ACM SIGKDD International Conference
  on Knowledge Discovery \& Data Mining}, pages 1064--1072, 2019.

\bibitem[\protect\citeauthoryear{Wu \bgroup \em et al.\egroup
  }{2016}]{wu2016methodology}
Xi~Wu, Matthew Fredrikson, Somesh Jha, and Jeffrey~F Naughton.
\newblock A methodology for formalizing model-inversion attacks.
\newblock In {\em 2016 IEEE 29th CSF}, pages 355--370. IEEE, 2016.

\bibitem[\protect\citeauthoryear{Wu \bgroup \em et al.\egroup
  }{2019a}]{wu2019adversarial}
Huijun Wu, Chen Wang, Yuriy Tyshetskiy, Andrew Docherty, Kai Lu, and Liming
  Zhu.
\newblock Adversarial examples on graph data: Deep insights into attack and
  defense.
\newblock {\em arXiv preprint arXiv:1903.01610}, 2019.

\bibitem[\protect\citeauthoryear{Wu \bgroup \em et al.\egroup
  }{2019b}]{wu2019session}
Shu Wu, Yuyuan Tang, Yanqiao Zhu, Liang Wang, Xing Xie, and Tieniu Tan.
\newblock Session-based recommendation with graph neural networks.
\newblock In {\em AAAI}, volume~33, pages 346--353, 2019.

\bibitem[\protect\citeauthoryear{Xu \bgroup \em et al.\egroup
  }{2019}]{xu2019topology}
Kaidi Xu, Hongge Chen, Sijia Liu, Pin-Yu Chen, Tsui-Wei Weng, Mingyi Hong, and
  Xue Lin.
\newblock Topology attack and defense for graph neural networks: An
  optimization perspective.
\newblock {\em IJCAI}, 2019.

\bibitem[\protect\citeauthoryear{Zhang \bgroup \em et al.\egroup
  }{2020}]{zhang2020secret}
Yuheng Zhang, Ruoxi Jia, Hengzhi Pei, Wenxiao Wang, Bo~Li, and Dawn Song.
\newblock The secret revealer: generative model-inversion attacks against deep
  neural networks.
\newblock In {\em CVPR}, pages 253--261, 2020.

\end{thebibliography}

\end{document}


\maketitle
\section{Dataset Statistics}
Here we list the statistics of 7 public dataset in Table \ref{data}.
\begin{table}[h]
\centering
\begin{tabular}{ccccc}
\toprule
         & Nodes & Edges & Classes & Features \\ \midrule
Cora     & 2,708  & 5,429  & 7       & 1,433     \\ 
Citeseer & 3,327  & 4,732  & 6       & 3,703     \\ 
Polblogs & 1,490  & 19,025  & 2       & -     \\ 
USA & 1,190  & 13,599  & 4       & -     \\
Brazil & 131  & 1,038  & 4       & -     \\
AIDS & 31,385  & 64,780  & 38       & 4     \\
ENZYMES & 19,580  & 74,564  & 3       & 18     \\
\bottomrule
\end{tabular}
\caption{Dataset statistics}
\label{data}
\end{table}
\section{Details of Random Sampling}

\begin{algorithm}[bh]
\caption{Random sampling from probabilistic vector to binary adjacency matrix}
\label{random sampling}
\textbf{Input}: Probabilistic vector $\mathbf{a}$, number of trials $K$\\
\textbf{Parameter}: Edge density $\rho$\\
\textbf{Output}: Binary matrix $A$;
\begin{algorithmic}[1] 
\STATE Normalize probabilistic vector: $\hat{\mathbf{a}}$ = $\mathbf{a}/\|\mathbf{a}\|_1$
\FOR{k = 1,2 $\cdots$ K}
Draw binary vector $\mathbf{a}^{(k)}$ by sampling $\lfloor \rho n \rfloor$ edges according to probabilistic vector $\hat{\mathbf{a}}$.
\ENDFOR
\STATE Choose a vector $\mathbf{a}^*$ from $\{\mathbf{a}^{(k)}\}$ which yields the smallest loss $\mathcal{L}_{attack}$. 
\STATE Convert $\mathbf{a}^*$ to binary adjacency matrix $A$ 
\STATE \textbf{return} $A$
\end{algorithmic}
\end{algorithm}

\section{Proof of Edge Influence}

\begin{theorem}
The adversary advantage is greater for edges with greater influence.
\end{theorem}

The proof is based on the lemma \ref{lemma} from \cite{wu2016methodology}.

\begin{Lemma}
\label{lemma}
Suppose the target model $f$ is trained on data distribution $p(\mathcal{X},\mathcal{Y})$. $\mathcal{X}= (x_s, x_{ns})$, where $x_s$ and $x_{ns}$ denote the sensitive and non-sensitive part of feature respectively.  The optimal adversary advantage is
\[P_{\mathcal{X} \sim p(\mathcal{X},\mathcal{Y})}[f(x_s=1, x_{ns}) \neq f(x_s=0, x_{ns})] .\]
\end{Lemma}

\noindent

\begin{proof}
In our graph setting, $x_{ns}$ refers to feature matrix $X$ and $x_s$ refers to edges.
Set $P[f^i_{\theta^*}(A, X)= y_i]$ = $P_{f_{\theta^*}}(y_i~ |~ A, X) = p$; $P[f^i_{\theta^*}(A_{-e}, X)= y_i] = P_{f_{\theta^*}}(y_i ~|~ A_{-e}, X) = q$.
Without loss of generality, the prediction accuracy is higher with edge $e$.
So, we set $q\le p$.
The adversary advantage is $Adv =[p(1-q)+q(1-p)]$.
Through variable substitution ($x=p-q, y=p+q$), we have $Adv= (y+\frac{x^2-y^2}{2}) $ and $\frac{\partial Adv}{\partial \mathcal{I}(e)} = \frac{\partial Adv}{\partial x}= p-q \ge 0$.
\end{proof}

\bibliographystyle{named}
\bibliography{paper}


\maketitle
\section{Summary Review}
The paper studies the problem of model inversion attack of GNNs, i.e., inferring the edges from GNNs. The reviewers generally agree the studied problem is interesting/novel and the use of projected gradient descent is feasible. The author rebuttal also clarified several concerns raised by reviewers, e.g., the generated graph is not an adversarial example and does not require to be unnoticeable. However, the novelty of the proposed attack is kind of limited from the methodological perspective. The authors are also encouraged to improve the paper presentation, provide comparisons with more state-of-the-art baselines, add convergence analysis, and provide justifications of the chosen datasets and assumptions of underlying attacks (e.g., the attacker's knowledge and capability).
\section{Changes of Resubmission Paper}
The authors are grataful for the valuable suggestions offered by reviewers. We have conducted major revisions to address the reviewers' concerns and improve this paper. Firstly, the authors provide comparisons with state-of-the-art baselines on more datasets. Secondly, convergence analysis, parameter settings, dataset separation and assumptions of the attacks are demonstrated clearly in the new paper. Thirdly, we have modified our model inversion method which takes the intrinsic properties of graph into consideration. Last but not least, we improved the presentation of the whole paper to make it easier for reviewers and readers to understand and follow.
\section{Reviews}
\subsection{Reviewer 1}
1. {Summary} Please summarize the main claims/contributions of the paper in your own words.\\
This paper proposes an interesting problem: Graph Model Inversion, that infers the edges from the known GNN models.\\
2. {Novelty} How novel is the paper?\\
Paper make non-trivial advances over past work\\
3. {Soundness} Is the paper technically sound?\\
I have not checked all details, but the paper appears to be technically sound\\
4. {Impact} How important is the paper likely to be, considering both methodological contributions and impact on application areas?\\
The paper will impact a moderate number of researchers\\
5. {Clarity} Is the paper well-organized and clearly written?\\
Good: paper is well organized but language can be improved\\
6. {Evaluation} Are claims well supported by experimental results?\\
Moderate: Experimental results are weak: important baselines are missing, or improvements are not significant\\
7. {Resources} How impactful will this work be via sharing datasets, code and/or other resources?\\
Not applicable: no shared resources\\
8. (Reproducibility) Would the experiments in the paper be easy to reproduce? (It may help to consult the paper’s reproducibility checklist.)\\
Good: e.g., code/data available, but some details of experimental settings are missing/unclear\\
\\
9. {Reasons to Accept} Please describe the paper’s key strengths.\\
1. Novel problem and intersting setting.\\
2. Using PGD to solve this problem\\
\\
10. {Reasons to Reject} Please describe the paper’s key weaknesses.\\
1. The motiveation to adopt PGD should be elaborated.\\
2. Comparsion to the other inversion attack is needed.\\
3. The explanation of the algorithm is not clear.\\
4. Experimental setting is node clear.\\
\\
11. {Detailed Comments} Please provide other detailed comments and constructive feedback.\\
In general, I found this work interesting, since the research on the problem of model inversion attack targeting graph neural networks is quite few and novel. However, the current manuscript can be further improved; see my questions/comments as below.\\
\\
1. I agree with the authors that projected gradient decent is used given the discrete nature of graphs as in [1]. However, could the authors provide a theoretical or empirical analysis on how the error would be if pure gradient decent is used under graph setting other than projected gradient decent.\\
\\
2. Further, though the authors mention that none of the existing model inversion attack methods can be directly used under the graph setting, it would be interesting to see how these existing model inversion attack would perform with the same setting. This could be implemented by such as naively adding threshold to make the recovered features in these methods discrete.\\
\\
3. In step 4 of Algorithm 1, the statement “if needed” is quite vague and should be explained more in details.\\
\\
4. Details upon how the baselines (embedding similarity and feature similarity) were constructed and tuned in your experimental settings are missed, such as how to select the threshold, since the embedding similarity also achieves comparable performance.\\
\\
------ Post Rebuttal -------\\
I've read the rebuttal from authors and most of my points has been addressed. Moreover, I think the novelty of the problem setting should be considered. Despite the negative opiontion from some reviewers, I would be positive to accept this paper.
\\
[1] Xu, K., Chen, H., Liu, S., Chen, P. Y., Weng, T. W., Hong, M., & Lin, X. (2019, August). Topology attack and defense for graph neural networks: an optimization perspective. In Proceedings of the 28th International Joint Conference on Artificial Intelligence (pp. 3961-3967). AAAI Press.\\
\\
12. {QUESTIONS FOR THE AUTHORS} Please provide questions for authors to address during the author feedback period. (Please number them)\\
See the detailed comments.\\
13. {Ethical Considerations} Please highlight any ethical considerations that must be considered before a final decision (it may help to consult the paper’s ethics statement, if provided)\\
None\\
\\
14. (OVERALL SCORE)\\
6 - Above threshold of acceptance\\
\\
19. I acknowledge that I have read the author's rebuttal and made whatever changes to my review where necessary.\\
Agreement accepted\\
\subsection{Reviewer 2}
1. {Summary} Please summarize the main claims/contributions of the paper in your own words. (Do not provide any review in this box)\\
The paper (1). designs a gradient based, white-box strategy, Graph Model Inversion (GMI), to generate the graph adversarial samples which are used to attack the SOTA graph models (e.g. GCN). (2). It also introduces the differential privacy method into graph model to help the graph model to defend the attack from the adversarial samples generated in (1).\\
2. {Novelty} How novel is the paper?\\
Paper contributes some new ideas\\
3. {Soundness} Is the paper technically sound?\\
I have not checked all details, but the paper appears to be technically sound\\
4. {Impact} How important is the paper likely to be, considering both\\ methodological contributions and impact on application areas?
The paper will impact a moderate number of researchers\\
5. {Clarity} Is the paper well-organized and clearly written?\\
Good: paper is well organized but language can be improved\\
6. {Evaluation} Are claims well supported by experimental results?\\
Moderate: Experimental results are weak: important baselines are missing, or improvements are not significant\\
7. {Resources} How impactful will this work be via sharing datasets, code and/or other resources? (It may help to consult the paper’s reproducibility checklist.)\\
Fair: some may find shared resources useful in future work\\
8. (Reproducibility) Would the experiments in the paper be easy to reproduce? (It may help to consult the paper’s reproducibility checklist.)
Meets Minimum Standard: e.g., code/data unavailable, but paper is clear enough that an expert could confidently reproduce\\
9. {Reasons to Accept} Please describe the paper’s key strengths.\\
(1). Besides proposing the attacking strategy, the paper also proposed a defense strategy to protect the graph model from the attack by the adversarial samples.\\
(2.) The paper analyzed the correlation between edge importance and inversion risk in a way.\\
10. {Reasons to Reject} Please describe the paper’s key weaknesses.\\
1. The proposed gradient based method for attacking GNN is not very novel.\\
2. The discussion of Experiment part is not very clear.\\
3. Many previous related works are not considered in the paper.\\
11. {Detailed Comments} Please provide other detailed comments and constructive feedback.\\
(1). One of an important thing in adversarial attack on graph model is that the perturbation on generated adversarial graph should be unnoticeable. Therefore, the paper should show some evidences (e.g. node degree distribution) to show the similarity between the original graphs and their adversarial samples.\\
\\
(2). In the Method part, I doubt that max L(a) may cause the asymmetry of adjacency matrix A.\\
\\
(3). As for the related work, the author ignores a lot of previous works. Many adversarial attack works on GNN have been proposed including meta-gradient on adjacency (discrete relaxation), black-box attack via reinforcement learning:\\
For example:\\
Adversarial attacks on neural networks for graph data, Daniel Zugner\\
12. {QUESTIONS FOR THE AUTHORS} Please provide questions for authors to address during the author feedback period. (Please number them)\\
(1). A few discussions in Experiment part are confused. For example: (1), The RQ1 want to show the effect of the designed attack strategy. However, the author used results in table 1 to support RQ1 which seems not valid. I think it may be solid to use the Figure 5 (a-b) to support the RQ 1. (2), when reading the Fig. 5 (a-b), why there are two “Projection lower bound”? One of them may be “upper bound”?\\
\\
(2). One of an important thing in adversarial attack on graph model is that the perturbation on generated adversarial graph should be unnoticeable. Therefore, the paper should show some evidences (e.g. node degree distribution) to show the similarity between the original graphs and their adversarial samples.\\
\\
14. (OVERALL SCORE)\\
5 - Below threshold of acceptance\\
\\
19. I acknowledge that I have read the author's rebuttal and made whatever changes to my review where necessary.\\
Agreement accepted
\subsection{Reviewer 3}
1. {Summary} Please summarize the main claims/contributions of the paper in your own words. (Do not provide any review in this box)\\
A model inversion attack on GNN based on the GNN model, and features and labels from a subset of vertices is proposed. The attack is based on solving a constrained optimization problem using PGD. A defense based on adding edges with high feature similarity is also proposed.\\
2. {Novelty} How novel is the paper?\\
Main ideas of the paper are known or incremental advances over past work\\
3. {Soundness} Is the paper technically sound?\\
I have not checked all details, but the paper appears to be technically sound\\
4. {Impact} How important is the paper likely to be, considering both methodological contributions and impact on application areas?\\
The paper will have low overall impact\\
5. {Clarity} Is the paper well-organized and clearly written?\\
Good: paper is well organized but language can be improved\\
6. {Evaluation} Are claims well supported by experimental results?\\
Poor: The experimental design is flawed and unconvincing\\
7. {Resources} How impactful will this work be via sharing datasets, code and/or other resources? (It may help to consult the paper’s reproducibility checklist.)\\
Fair: some may find shared resources useful in future work\\
8. (Reproducibility) Would the experiments in the paper be easy to reproduce? (It may help to consult the paper’s reproducibility checklist.)\\
Meets Minimum Standard: e.g., code/data unavailable, but paper is clear enough that an expert could confidently reproduce\\
9. {Reasons to Accept} Please describe the paper’s key strengths.\\
The paper addresses a timely and important topic on privacy of the connectivity graph in GNN models.\\
10. {Reasons to Reject} Please describe the paper’s key weaknesses.\\
The experiment results are unconvincing due to the datasets used. Unclear what is the performance in bigger and more diverse datasets.\\
11. {Detailed Comments} Please provide other detailed comments and constructive feedback.\\
1. Please state the dimensions of each parameter like $W^k$ carefully. $M_k$ is used in (1) and also as the dimension of the feature of node v in layer k. The equation (3) does not seem to be properly written.\\
\\
2. Please proofread the paper. There are various grammatical mistakes. E.g., "one optimization problem" should be "an optimization problem", "will severely undermines", "the rest missing links", etc. There are also formatting mistakes like writing "if" in math font instead of text font.\\
\\
12. {QUESTIONS FOR THE AUTHORS} Please provide questions for authors to address during the author feedback period. (Please number them)\\
1. How large is $V_GMI$? There should be investigations to show the impact of the subset size and choice on the attack and defense performance.\\
\\

2. How important are labels in the attack performance? Why are labels not considered in the defense mechanism? [My point was why the authors did not consider using labels in the defense mechanism and why they designed their defense as such? They have not answered this question satisfactorily.]\\
\\
3. It is well known that the inference performance of the datasets Cora and Citeseer are highly sensitive to the underlying graph topology (after all, these are citation networks based on word documents). Hence there is no surprise in Fig. 4. To test the efficacy of the proposed attacks, the authors should consider other datasets where node features play a more important role. [No satisfactory answer from authors.]\\
\\

4. Unclear if the paper is performing transductive learning of the GNN model. Are the GCN, GAT and GraphSAGE pre-trained models or the authors retrain them? How are the testing nodes chosen?\\
13. {Ethical Considerations} Please highlight any ethical considerations that must be considered before a final decision (it may help to consult the paper’s ethics statement, if provided)
\\
Nil\\
14. (OVERALL SCORE)\\
3 - Clear reject\\
19. I acknowledge that I have read the author's rebuttal and made whatever changes to my review where necessary.\\
Agreement accepted
\subsection{Reviewer 4}
Questions
1. {Summary} Please summarize the main claims/contributions of the paper in your own words. (Do not provide any review in this box)\\
This paper studies an important problem, i.e., model inversion attack of Graph Neural Networks (GNN) models. Due to the discreteness of graph data, the authors resort to projected gradient descent with convex relaxation and design a denoising module to improve the performance of model inversion attack on GNNs. The authors also provide studies on the connection between model inversion risk and edge importance and find edges with higher importance are more likely to be recovered, and propose a simple defense method based on graph preprocessing. Furthermore, the authors also conduct experiments to evaluate the effectiveness of the proposed methods.\\
2. {Novelty} How novel is the paper?\\
Paper contributes some new ideas\\
3. {Soundness} Is the paper technically sound?\\
The paper has minor technical flaws that are easily fixable\\
4. {Impact} How important is the paper likely to be, considering both methodological contributions and impact on application areas?\\
The paper will impact a moderate number of researchers\\
5. {Clarity} Is the paper well-organized and clearly written?\\
Good: paper is well organized but language can be improved\\
6. {Evaluation} Are claims well supported by experimental results?\\
Good: Experimental results are sufficient, though more analysis would significantly add support to the claims\\
7. {Resources} How impactful will this work be via sharing datasets, code and/or other resources? (It may help to consult the paper’s reproducibility checklist.)\\
Fair: some may find shared resources useful in future work\\
8. (Reproducibility) Would the experiments in the paper be easy to reproduce? (It may help to consult the paper’s reproducibility checklist.)\\
Good: e.g., code/data available, but some details of experimental settings are missing/unclear\\
9. {Reasons to Accept} Please describe the paper’s key strengths.\\
$[+]$ The authors propose graph model inversion attack which applies projected gradient descent on graphs via convex relaxation for edge reconstruction.\\
\\
$[+]$ The authors evaluate the relation between edge importance and model inversion risk and find edges with higher importance are more likely to be reconstructed.\\
\\
$[+]$ The authors propose one defense method.\\
10. {Reasons to Reject} Please describe the paper’s key weaknesses.\\
$[-]$ It would be better if the authors empirically and theoretically analyze the convergence of the proposed method.\\
\\
$[-]$ The authors fail to cite state-of the-art works. Currently, different defense mechanisms against model inversion attacks are proposed, e.g., [1,2,3]. It would be better if the authors give more discussion and conduct comparative experiments.\\
\\
$[-]$ The assumptions about attacker’s knowledge and capability that the authors make are very strong.\\
\\
$[-]$ More details are needed. The authors relax $a$ into convex space. It would be better if the authors analyze the approximation errors before and after the relaxation of the vector $a$. Additionally, it would be better if the authors provide more details on how to select the optimal dropout rates, which are applied to adjacency vector $a$.\\
\\
$[1]$ “Improving Robustness to Model Inversion Attacks via Mutual Information Regularization”, 2020.\\
$[2]$ “Defending Model Inversion and Membership Inference Attacks via Prediction Purification”, 2020.\\
$[3]$ “Model inversion attacks that exploit confidence information and basic countermeasures”, 2015.\\
\\
11. {Detailed Comments} Please provide other detailed comments and constructive feedback.\\
This paper studies an important problem, i.e., model inversion attack of Graph Neural Networks (GNN) models. Due to the discreteness of graph data, the authors resort to projected gradient descent with convex relaxation and design a denoising module to improve the performance of model inversion attack on GNNs. The authors also provide studies on the connection between model inversion risk and edge importance and find edges with higher importance are more likely to be recovered, and propose a simple defense method based on graph preprocessing. Furthermore, the authors also conduct experiments to evaluate the effectiveness of the proposed methods. However, I still have the following concerns.
\\
$[-]$ It would be better if the authors empirically and theoretically analyze the convergence of the proposed method.\\
\\
$[-]$ The authors fail to cite state-of the-art works. Currently, different defense mechanisms against model inversion attacks are proposed, e.g., [1,2,3]. It would be better if the authors give more discussion and conduct comparative experiments.\\
\\
$[-]$ The assumptions about attacker’s knowledge and capability that the authors make are very strong.\\
\\
$[-]$ More details are needed. The authors relax $a$ into convex space. It would be better if the authors analyze the approximation errors before and after the relaxation of vector $a$. Additionally, it would be better if the authors provide more details on how to select the optimal dropout rates, which are applied to adjacency vector $a$.\\
\\
$[1]$ “Improving Robustness to Model Inversion Attacks via Mutual Information Regularization”, 2020.\\
$[2]$ “Defending Model Inversion and Membership Inference Attacks via Prediction Purification”, 2020.\\
$[3]$ “Model inversion attacks that exploit confidence information and basic countermeasures”, 2015.\\
\\
12. {QUESTIONS FOR THE AUTHORS} Please provide questions for authors to address during the author feedback period. (Please number them)
Please see the above detailed comments.\\
14. (OVERALL SCORE)\\
5 - Below threshold of acceptance\\
19. I acknowledge that I have read the author's rebuttal and made whatever changes to my review where necessary.\\
Agreement accepted\\